\documentclass[letterpaper]{article} 
\usepackage{aaai24}  % DO NOT CHANGE THIS
\usepackage{times}  % DO NOT CHANGE THIS
\usepackage{helvet}  % DO NOT CHANGE THIS
\usepackage{courier}  % DO NOT CHANGE THIS
\usepackage[hyphens]{url}  % DO NOT CHANGE THIS
\usepackage{graphicx} % DO NOT CHANGE THIS
\usepackage{natbib}  % DO NOT CHANGE THIS AND DO NOT ADD ANY OPTIONS TO IT
\usepackage{caption} % DO NOT CHANGE THIS AND DO NOT ADD ANY OPTIONS TO IT
\usepackage{amsthm}
\usepackage{subfigure}
\usepackage{amsmath}
\usepackage{amssymb}
\usepackage{multirow}
\usepackage{caption}
\usepackage{mathrsfs}

\usepackage{colortbl}
\usepackage{mathtools}

\usepackage{lipsum,booktabs}

\theoremstyle{plain}
\newtheorem{theorem}{Theorem}[section]
\newtheorem{proposition}[theorem]{Proposition}

\theoremstyle{definition}
\newtheorem{definition}[theorem]{Definition}

\theoremstyle{remark}
\newtheorem{remark}[theorem]{Remark}

\usepackage{newfloat}
\usepackage{listings}
\title{TopoGCL: Topological Graph Contrastive Learning}
\author {
    Yuzhou Chen\textsuperscript{\rm 1},
    Jose Frias\textsuperscript{\rm 2},
    Yulia R. Gel\textsuperscript{\rm 3,4}\\
}
\affiliations {
   \textsuperscript{\rm 1}Department of Computer and Information Sciences, Temple University\\
   \textsuperscript{\rm 2}Department of Mathematics, UNAM\\
\textsuperscript{\rm 3}Department of Mathematical Sciences, University of Texas at Dallas\\
     \textsuperscript{\rm 4}National Science Foundation\\
yuzhou.chen@temple.edu, frias@ciencias.unam.mx, ygl@utdallas.edu
}

\usepackage{bibentry}
\begin{document}

\maketitle

\begin{abstract}
Graph contrastive learning (GCL) 
has recently emerged as a new concept which allows for capitalizing on the  strengths of graph neural networks (GNNs) to learn rich representations in a wide variety of applications which involve abundant unlabeled information.
However, existing GCL approaches largely tend to overlook the important latent information on higher-order graph substructures. We address this limitation by introducing the concepts of topological invariance and extended persistence on graphs to GCL. In particular, we propose a new contrastive mode which targets topological representations of the two augmented views from the same graph, yielded by extracting latent shape properties of the graph at multiple resolutions. Along with the extended topological layer, we introduce a new extended persistence summary, namely, extended persistence landscapes (EPL) and derive its theoretical stability guarantees. Our extensive numerical results on biological, chemical, and social interaction graphs show that the new Topological Graph Contrastive Learning (TopoGCL) model delivers 
significant performance gains in
 unsupervised graph classification for 11 out of 12 considered datasets and also exhibits robustness under noisy scenarios. 
\end{abstract}

\section{Introduction}

In the last couple of years self-supervised contrastive learning (CL) has emerged as a new promising trend in graph learning which has brought the power of Graph Neural Networks (GNNs) to a broad range of applications without annotated supervisory data, from prediction of molecular properties in biochemistry to discovery of new crystalline materials~\cite{koker2022graph, xu2021self,fang2022molecular,stark20223d}. 
Indeed, until very recently GNNs have been limited in their 
representation learning capabilities due to the over-reliance on the existence of task-dependent labels.
However, such supervisory information is often hand-crafted and may be both scarce and notoriously hard to obtain in many real-life applications of graph learning. For instance, labeling e-crime activity  on blockchain transaction graphs typically involves a highly resource-intensive process of manual annotation by the law enforcement agencies, while in bioinformatics and material research graph labeling requires costly and time-consuming wet-lab experiments. 

The emerging GCL paradigm rests on the two major components: 1) {\it augmentation} which constructs multiple views of the graph by exploiting invariance under various transformations such as subgraph sampling, perturbations of nodes and edges, and attribute masking~\cite{you2020graph,zeng2021contrastive}; and 2) {\it contrastive learning (CL)} itself which  
maximizes mutual information (MI) among the views generated from the resulting graph augmentations, such that positive pairs are contrasted with their negative counterparts~\cite{sun2019infograph,velickovic2019deep,hassani2020contrastive}. The CL step is performed by contrasting node-level representations and graph-level representations, and the three traditional contrasting modes are local-local CL, global-local CL, and global-global CL. However, since the agreement analysis among representations is typically assessed using cosine similarity of the related embeddings,
these contrasting approaches cannot systematically account for similarity
of higher-order graph properties, for instance, simultaneous matching among subgraphs of varying sizes and orders. In turn, such polyadic node interactions, including various network motifs and other multi-node graph substructures, often play the key role in graph learning tasks, especially, in conjunction with prediction of protein functions in protein-protein interactions and fraud detection in financial networks~\cite{benson2016higher,chen2022bscnets}.
Interestingly, as shown by~\cite{you2020graph}, subgraphs also tend to play the uniformly consistent role in the data augmentation step of GCL across all types of the considered graphs, from bioinformatics to social networks.

Motivated by the recent success of the 
computational algebraic topology in graph representation learning, we introduce the concepts of  topological representation invariance and extended persistent homology (PH) to GCL.
In particular, PH is a tool in computational topology which retrieves evolution of the shape patterns in the observed data along various user-defined geometric dimensions~\cite{hofer2019learning,edelsbrunner2000topological,zomorodian2005computing}. By ``shape'' here, we broadly refer to the 
data properties which are invariant  under continuous transformations such as bending,  stretching, and twisting.
PH tools, often in a form of a fully trainable topological layer~\cite{carriere2020perslay,chen2021topological,yan2021link,horntopological}, have proliferated into a wide range of graph learning tasks, from node and graph classification to anomaly detection to link prediction, often resulting not only in bolstering GNN performance but also in enhancing robustness against random perturbations and adversarial attacks. 
Nevertheless, despite the well-documented PH utility in semi-supervised graph learning and the intuitive premise of
topological invariance to better assess similarity among graph augmentations (see a visual experiment with proteins in Figure~\ref{epd_proteins_plot} and Appendix~D.2), PH concepts have never been explored in conjunction with GCL. Here we bridge this gap and introduce the new contrastive mode which targets topological representations of the two augmented views from the same graph. We specifically focus on the notion of extended persistence (EP) which, despite being under-explored in ML applications, has been shown to retrieve a richer topological structure from the observed data and, hence, be particularly suitable for shape matching within GCL. Furthermore, we introduce a new EP summary, extended persistence landscapes (EPL) and prove its stability guarantees. We also contrast  theoretical and numerical properties of EPL with respect to another EP summary, extended persistence images (EPI). While EPI has been used in applications before~\cite{yan2021link}, it has neither been formally defined for the EP case nor its properties have been evaluated. Armed with EPL, EPI and the associated extended topological layer, we develop a new Topological Graph Contrastive Learning (TopoGCL) model, equipped with contrastive mode on extended topological representations (topological-topological CL ({\it aka} topo-topo CL)) that allows us to capture not only the critical topological and geometric graph information but also to enhance its latent representation learning.

Significance of our contributions can be summarized as follows:
\begin{itemize}
\item TopoGCL is the first approach introducing the concepts of persistent homology to graph contrastive learning.

\item We propose a new summary of extended persistence, namely, extended persistence landscapes, and prove its theoretical stability guarantees.

\item We validate the utility of TopoGCL in conjunction with
unsupervised graph classifications on 12 benchmark datasets from biology, chemistry, and social sciences. Our findings indicate that in addition to outperforming state-of-the-art baselines on 11 out of 12 benchmarks and delivering (statistically) significant gains on 8 datasets, TopoGCL also yields highly promising results in terms of robustness to noise.

\end{itemize}

%such as graph topological information from data to learn comprehensive embeddings.

%Typically, CL is capable to learn the discriminative embedding from node-level or graph-level, and the traditional contrasting modes including local-local CL, global-local CL, and global-global CL. 

% Yulia, how about that.(let me add it.) Thanks a lot! But what are the drawbacks? Lack of higher informarion

%\YGL{Yuzhou, can you please write on the existing approaches?} 

\section{Related Work}
{\bf Graph Representation Learning and Graph Contrastive Learning.}
Recently, inspired by the success of convolutional neural networks (CNN) on image-based tasks, graph neural networks (GNNs) have emerged as a powerful tool for graph representation learning. Based on the spectral graph theory,~\cite{bruna2013spectral} introduces a graph-based convolution in Fourier domain. However, the complexity of this model is very high since all Laplacian eigenvectors are needed. To tackle this problem, ChebNet~\cite{defferrard2016convolutional} integrates spectral graph convolution with Chebyshev polynomials. Then, Graph Convolutional Networks (GCNs) of~\cite{kipf2016semi} simplify the graph convolution with a localized first-order approximation. More recently, there have been proposed various approaches based on accumulation of the graph information from a wider neighborhood, using diffusion aggregation and random walks. Such higher-order methods include approximate personalized propagation of neural predictions (APPNP)~\cite{klicpera2018predict}, and higher-order graph convolutional architectures (MixHop)~\cite{abu2019mixhop}. Moreover, other recent approaches include GNNs with the attention mechanism (GAT, SPAGAN)~\cite{velivckovicgraph,yang2019spagan}, and GNNs based on graph diffusion convolution~\cite{gasteiger2019diffusion, zhao2021adaptive}. Furthermore, there has appeared a number of approaches introducing a pooling mechanism into GNNs to capture graph (sub)structural information~\cite{cangea2018towards,gao2019graph,lee2019self, du2021multi}. However, such tools mainly focus on supervised and semi-supervised settings and differ from our unsupervised representation learning scheme. Graph contrastive learning is a self-supervised learning approach to learn an encoder (e.g., GNNs without the final classifier) for extracting embeddings from the unlabeled input data. Existing graph contrastive learning approaches mainly focus on three modes, i.e., local-local CL~\cite{zhu2021graph,thakoor2021bootstrapped}, global-local CL~\cite{velickovic2019deep, sun2019infograph}, and global-global CL~\cite{you2020graph,li2022let}. For instance, GCC~\cite{qiu2020gcc} proposes a pretraining framework based on local-local CL which constructs multiple graph
views by sampling subgraphs based on random walks. For global-local CL, the works~\cite{velickovic2019deep, hassani2020contrastive, asanoself, hassani2020contrastive} follow the InfoMax principle~\cite{linsker1988self} to maximize the Mutual Information (MI) between the representation of local features and global features. Moreover, another graph contrastive learning mode, i.e., global-global CL~\cite{you2020graph, fang2022molecular} studies the relationships between the global context representations of different samples, which performs better on graph-level tasks. Different from these methods, we propose a novel model Topological Graph Contrastive Learning with the topo-topo CL contrasting mode that not only captures crucial topological and geometrical information but enhances the latent graph representation learning.

{\bf Extended Persistence for Machine Learning.} Extended persistence (EP) has been introduced by~\cite{cohen2009extending} who show that, by assessing the evolution of shape properties in both upward and downward filtration direction, EP allows us to capture some important topological properties of the underlying object that ordinary persistence cannot. This makes EP particularly attractive for shape matching and CL. However, EP remains substantially less
explored in the ML literature, comparing with the ordinary persistence~\cite{carlsson2021topological, adams2021topology, pun2022persistent}. Some prominent applications of EP in graph learning include link prediction~\cite{yan2021link}, node classification~\cite{zhao2020persistence,yan2022neural}, and graph classification~\cite{carriere2020perslay}. To our best knowledge neither EP, nor ordinary persistence or any other tools of computational topology have been ever applied in conjunction with contrastive learning. TopoGCL is the first approach to bridge this knowledge gap.

%\section{Methodology}
\section{Notations and Preliminaries}
Let $\mathcal{G} = (\mathcal{V}, \mathcal{E}, \boldsymbol{X})$ be an attributed graph, where $\mathcal{V}$ is a set of nodes ($|\mathcal{V}|=N$), $\mathcal{E}$ is a set of edges, and $\boldsymbol{X} \in \mathbb{R}^{N \times F}$ is a node feature matrix (here $F$ is the dimension of node features). Let $\boldsymbol{A} \in \mathbb{R}^{N \times N}$ be a symmetric adjacency matrix such that
$\boldsymbol{A}_{uv} = \omega_{uv}$ if nodes $u$ and $v$ are connected and 0, otherwise 
(here $\omega_{uv}$ is an edge weight and $\omega_{uv}\equiv 1$ for unweighted graphs).
Furthermore, $\boldsymbol{D}$ represents the degree matrix with $\boldsymbol{D}_{uu} = \sum_v \boldsymbol{A}_{uv}$, corresponding to $\boldsymbol{A}$. 

{\bf Preliminaries on Extended Persistent Homology.}
PH is a subfield in computational topology whose main goal is to detect, track and encode the evolution of shape patterns in the observed object along various user-selected geometric dimensions~\cite{edelsbrunner2000topological,zomorodian2005computing, carlsson2021topological}.
These shape patterns represent topological properties such as  connected components, loops, and, in general, $n$-dimensional ``holes", that is, the characteristics of the graph $\mathcal{G}$ that remain preserved at different resolutions under continuous transformations. By employing such a multi-resolution approach, PH addresses the intrinsic limitations of classical homology and allows for retrieving the latent shape properties of $\mathcal{G}$ which may play the essential role in a given learning task. The key approach here is to select some suitable scale parameters $\nu$ and then to study changes in shape of $\mathcal{G}$ that occur as $\mathcal{G}$ evolves with respect to $\nu$. That is, we no longer study $\mathcal{G}$ as a single object but as a {\it filtration} $\mathcal{G}_{\nu_1} \subseteq \ldots \subseteq \mathcal{G}_{\nu_n}=\mathcal{G}$, induced by monotonic changes of $\nu$. To ensure that the process of pattern selection and counting is objective and efficient, we build an abstract simplicial complex $\mathscr{K}(\mathcal{G}_{\nu_j})$ on each $\mathcal{G}_{\nu_j}$, which results in a filtration of complexes $\mathscr{K}(\mathcal{G}_{\nu_1}) \subseteq \ldots \subseteq \mathscr{K}(\mathcal{G}_{\nu_n})$. 
For example, for an edge-weighted graph $(\mathcal{V}, \mathcal{E}, w)$, with the edge-weight function $w: \mathcal{E} \rightarrow \mathbb{R}$, we can set $\mathcal{G}_{\leq\nu_j}=(\mathcal{V}, \mathcal{E}, w^{-1}(-\infty, \nu_j])$ for each $\nu_j$, $j=1,\ldots, n$, yielding the induced sublevel edge-weighted filtration. 
Similarly, we can consider a function on a node set $\mathcal{V}$, for example, node degree, which results in a sequence of induced subgraphs of $\mathcal{G}$ with a maximal degree of $\nu_j$ for each $j=1,\ldots, n$ and the associated degree sublevel set filtration. We can then record scales $b_i$ (birth) and $d_i$ (death) at which each topological feature first and last appear in the sublevel filtration
$\mathcal{G}_{\nu_1} \subseteq \mathcal{G}_{\nu_2} \subseteq \mathcal{G}_{\nu_3} \ldots$. However, in such sublevel filtration, some topological features may never disappear (i.e., persist forever), resulting in a loss of the important information on the underlying latent topological properties of $\mathcal{G}$ and, hence, making it more difficult to use the extracted topological information for shape matching among objects.
An alternative approach is to complement the sublevel filtration by its superlevel counterpart, that is, to consider also the sequence of superlevel subgraphs $\mathcal{G}^{\leq\nu_j}=(\mathcal{V}, \mathcal{E}, w^{-1}[\nu_j, \infty))$ for each $\nu_j$, $j=1,\ldots, n$. That is, we know
simultaneously assess evolution of topological features of $\mathcal{G}$ observed over filtrations in both upward and downward directions.  This mechanism results in the {\it extended} persistence, which encompassed information obtained both from sublevel and superlevel filtrations, and the death times can be now also defined as indices at which the topological feature reappears in the superlevel sequence of graphs $\mathcal{G}^{\leq\nu_j}$.   The extracted extended topological information can be then summarized as a multiset in $\mathbb{R}$ called {\it extended persistence diagram (EPD)} $\mathcal{\text{EDg}}=\{(b_{\rho},d_{\rho}) \in \mathbb{R}^2\}$. (For further details, see Appendix~C.) Note that the extended persistence ensures that no topological feature persist forever and is hence particularly suitable for latent shape matching, opening new perspectives for topological contrastive learning.

%\medskip
{\bf What New Does Topological Invariance Bring to GCL?} Figure~\ref{epd_proteins_plot} in Appendix D.2 shows 4 networks PROTEINS dataset, along with their corresponding EPDs. These protein networks are hard to discern visually and their traditional network summaries are also virtually indistinguishable. 
Furthermore, the state-of-the-art CL models also do not correctly classify these 4 proteins. However, we find that the Wasserstein distances between two EPDs of protein networks are always very high if the two protein networks do not belong to the same class and low, otherwise. This phenomenon underlines that persistence and topological invariance play important roles in CL. (For more details see Appendix~D.2.)

\section{Topological Graph Contrastive Learning}
We now introduce our topological graph contrastive learning (TopoGCL) model which incorporates both graph and topological representations learning into the CL module. In this section, we first briefly recap graph contrastive learning (GCL). Then we discuss the details of the proposed topological contrastive learning (TopoCL). The overall architecture is demonstrated in Figure~\ref{flowchart}. To facilitate the reading, the frequently used mathematical notations are summarized in Table~\ref{notations} in Appendix~A.1.

\subsection{Graph Contrastive Learning}
Consider a set of graphs $\boldsymbol{\mathcal{G}} = \{\mathcal{G}_1, \dots, \mathcal{G}_{\Upsilon}\}$. Following the InfoMax principle~\cite{linsker1988self}, GCL aims to perform pre-training through maximizing the mutual information between two augmented views of the same graph via a contrastive loss in the learned latent space~\cite{hassani2020contrastive, you2020graph}. For a better illustration, let us start, as an example, with a case of one graph $\mathcal{G}_i$, where $i \in \{1, \dots, \Upsilon\}$. That is, given $\mathcal{G}_i = \{\mathcal{V}_i, \mathcal{E}_i, \boldsymbol{X}_i\}$, we first corrupt the original $\mathcal{G}_i$ by an explicit corruption pipeline $\mathcal{T}(\cdot)$ (e.g., node perturbation, edge perturbation, or node feature shuffling) to convert the graph into the two perturbed versions $\Tilde{\mathcal{G}}_i = \mathcal{T}_i(\mathcal{G}_i) =\{\Tilde{\mathcal{V}}_i, \Tilde{\mathcal{E}}_i, \Tilde{\boldsymbol{X}}_i\}$ and $\Tilde{\mathcal{G}}^{\prime}_i = \mathcal{T}^\prime_i(\mathcal{G}_i) = \{\Tilde{\mathcal{V}}^{\prime}_i, \Tilde{\mathcal{E}}^{\prime}_i, \Tilde{\boldsymbol{X}}^{\prime}_i\}$. Then, both $\Tilde{\mathcal{G}_i}$ and $\Tilde{\mathcal{G}}^{\prime}_i$ are fed into a shared $f_\text{ENCODER}(\cdot)$ for graph representation learning (see the {\color{black}{blue}} box in Figure~\ref{flowchart}). Here $\Tilde{\boldsymbol{H}}_i = f_\text{ENCODER}(\Tilde{\mathcal{G}}_i)$ and $\Tilde{\boldsymbol{H}}^\prime_i = f_\text{ENCODER}(\Tilde{\mathcal{G}}^\prime_i)$ are the learned representations of the two augmented views of the original graph $\mathcal{G}_i$. The contrastive loss function for the positive pair of samples $\ell_{\text{G}}(\Tilde{\mathcal{G}}_i, \Tilde{\mathcal{G}}^\prime_i)$ is formulated as: 
\begin{align}
\label{graph_loss}
    \ell_{i, \text{G}}(\Tilde{\mathcal{G}}_i, \Tilde{\mathcal{G}}^\prime_i) = -\log\frac{\exp{(\text{sim}(\Tilde{\boldsymbol{H}}_{i}, \Tilde{\boldsymbol{H}}^\prime_{i})/\zeta)}}{\sum^{2\Upsilon}_{\gamma, \gamma \neq i} \exp{(\text{sim}(\Tilde{\boldsymbol{H}}_{i}, \Tilde{\boldsymbol{H}}_{\gamma})/\zeta)}},
\end{align}
where $\text{sim}(\Tilde{\boldsymbol{H}}_{i}, \Tilde{\boldsymbol{H}}^\prime_{i}) = \Tilde{\boldsymbol{H}}^{\top}_{i}\Tilde{\boldsymbol{H}}^\prime_{i}/||\Tilde{\boldsymbol{H}}_{i}|| ||\Tilde{\boldsymbol{H}}^\prime_{i}||$, $\Tilde{\boldsymbol{H}}_{\gamma} = f_{\text{ENCODER}}(\Tilde{\mathcal{G}}_{\gamma})$ denotes the graph representation of $\Tilde{\mathcal{G}}_{\gamma}$, and $\zeta$ is the temperature hyperparameter.

\subsection{Topological Contrastive Learning}

%The ultimate goal of topological contrastive learning (TopoCL) is to learn the topological representation invariant to specialized perturbations through a new contrasting model, i.e., {Topological-Topological CL (Topo-Topo CL)}. In summary, topo-topo CL targets at contrasting between topological representations of the two augmented views from the same graph. %More specifically, given a graph $\mathcal{G} = \{\mathcal{V}, \mathcal{E}, \boldsymbol{X}\}$, we first corrupt the original $\mathcal{G}$ by an explicit corruption pipeline (e.g., node perturbation, edge perturbation, and node feature shuffling) to convert the graph into two perturbed versions $\Tilde{\mathcal{G}}_i = \{\Tilde{\mathcal{V}}_i, \Tilde{\mathcal{E}}_i, \Tilde{\boldsymbol{X}}_i\}$ and $\Tilde{\mathcal{G}}_j = \{\Tilde{\mathcal{V}}_j, \Tilde{\mathcal{E}}_j, \Tilde{\boldsymbol{X}}_j\}$; next, by using extended persistent homology, we can obtain extended topological features from two perturbed graphs $\Tilde{\mathcal{G}}_i$ and $\Tilde{\mathcal{G}}_j$, respectively; finally, we apply our proposed topological layer to obtain latent extended topological embeddings, and contrast postive and negative topological embeddings at the same scale.

The ultimate goal of TopoCL is to contrast latent shape properties of the two augmented views from the same graph, assessed at different resolution scales, by contrasting their respective extended topological representations.
Below we introduce the two key components of our method, i.e., extraction of the extended topological features and extended topological representation learning. We focus our discussion on a perturbed graph $\Tilde{\mathcal{G}}$ for the sake of simplicity (omitting the subscript $i$).
%As aforementioned, there are multiple ways that a graph filtration can be constructed (see, e.g.,~\cite{hofer2020graph}). For instance, here we consider a sublevel filtration function based a centrality score $\mathfrak{F}$ defined on nodes of $\Tilde{\mathcal{G}}$. That is, let $\mathfrak{F}:\mathcal{V} \rightarrow \mathbb{R}$ and $\nu_1<\nu_2<\cdots <\nu_n$ be a sequence of sorted filtered values, then $\mathscr{K}_{i}=\{\sigma \in \mathscr{K}: \max_{v\in \sigma}\mathfrak{F}(v)\leq \nu_i\}$. (A filtration on edges of $\mathcal{G}$ can be defined in a similar manner.) Given the filtration $\mathfrak{F}$, we can obtain a $p$-dimensional extended persistence diagrams, i.e., $\Tilde{\text{EDg}}^{(p)}$ (where $p = 0, 1, \dots$). Note that, for learning graph topology, in our study, we only consider $p \in \{0, 1\}$, i.e., capturing connected components ($p = 0$) or cycles ($p=1$). 
Since an extended persistence diagram ${\text{EDg}}$ is a multiset in $\mathbb{R}^2$, it cannot be directly used as an input into ML models. %which is difficult to be used as inputs for ML and DL methods (due to the complicated space structure, cardinality issues, computationally inefficient metrics, etc.). 
To facilitate effective and flexible downstream applications, we utilize the representations of EPD in a functional Hilbert space, i.e., the extended persistence landscape (EPL) and the extended persistence image (EPI). We also design a novel extended topological layer based on the given ${\text{EDg}}$. As such, broadly speaking, TopoCL consists of three steps: 
%the construction of our proposed extended topological layer consists of two steps: 
(i) extracting the latent shape properties of the graph using extended persistence in a form of extended persistence diagram ${\text{EDg}}$ and then converting  ${\text{EDg}}$ into either EPL or EPI, (ii) constructing the extended topological layer, and (iii) then contrasting the derived topological representations.

{\bf Extended Persistence Vectorization.} Here we focus on two summaries for EP, Extended Persistence Landscapes and Extended Persistence Images. Both of them are motivated by their respective counterparts as summaries of ordinary persistence. However, while EPI has appeared in applications before~\cite{yan2021link}, it has not been formally defined. EPL is a new summary, and we derive its theoretical stability guarantees and discuss its properties in comparison with EPI.

%Inspired by persistence landscapes for ordinary persistence~\citet{Bubenik_2015}, we propose a new computationally efficient EP summary called Extended Persistence Landscape, derive its theoretical stability guarantees and discuss its properties in comparison with the extended persistence image which has been used in applications~\cite{yan2021link} but has not been formally defined.
%{$\color{gray}{\Tilde{\text{EDg}}} \rightarrow$ : 
{\bf Extended Persistence Landscape (EPL).} % $\color{gray}{{\text{EDg}}} \rightarrow{{\text{EPL}}}$}
%\YGL{Jose, please, clean this section, i.e., definition shall be formalized. Also please add some theoretical properties of EPL, even in a skeleton form.}
%Persistence Landscapes $\lambda$ are one of the most common persistence vectorizations introduced by~\cite{Bubenik_2015}. 
 Inspired by persistence landscapes for ordinary persistence of~\cite{Bubenik_2015}, we propose a new computationally efficient EP summary called Extended Persistence Landscape.
Consider the
generating functions $\Lambda_i$ for each $(b_i,d_i)\in {\text{EDg}}$, i.e., $\Lambda_i: \mathbb{R} \to \mathbb{R}$ is the 
piecewise linear function obtained by two line segments starting from $(b_i,0)$ and $(d_i,0)$ connecting to the same point $(\frac{b_i+d_i}{2},\frac{d_i-b_i}{2})$ and $0$ in $\mathbb{R}\setminus [b_i,d_i]$. Please see Appendix~C for an extended exposition on Extended Homology. 

%piece-wise linear functions obtained by the two-line segment from the point $(b_i,0)$ and to the point $(d_i,0)$, connecting to the same point $(\frac{b_i+d_i}{2},\frac{b_i-d_i}{2})$.
%For a given extended persistence diagram ${\text{EDg}}=\{(b_i,d_i)\}$,  $\lambda$ produces a function $\lambda({\mathcal{G}})$ by using generating functions $\Lambda_i$ for each $(b_i,d_i)\in \Tilde{\text{EDg}}$, i.e., $\Lambda_i:[b_i,d_i]\to \mathbb{R}$ is a piecewise linear function obtained by two line segments starting from $(b_i,0)$ and $(d_i,0)$ connecting to the same point $(\frac{b_i+d_i}{2},\frac{b_i-d_i}{2})$. 
%Persistence Landscapes $\lambda$ are one of the most common vectorizations of ordinary persistence introduced by~\cite{Bubenik_2015}. 

%For a given extended persistence diagram $\text{EDg}=\{(b_i,d_i)\}$, $\lambda$ produces a set of functions $\{\lambda_j(\Tilde{\mathcal{G}}):\mathbb{R} \rightarrow \mathbb{R}\}$ by using generating functions $\Lambda_i$ for each $(b_i,d_i)\in {\text{EDg}}$, i.e., $\Lambda_i:\mathbb{R}\to \mathbb{R}$ is a piecewise linear function defined in $[b_i,d_i]$ by two line segments starting from $(b_i,0)$ and $(d_i,0)$ connecting to the same point $(\frac{b_i+d_i}{2},\frac{d_i-b_i}{2})$ and defined as $0$ in $\mathbb{R}\setminus [b_i,d_i]$.  Inspired by persistence landscapes for ordinary persistence of~\cite{Bubenik_2015}, we propose a new computationally efficient EP summary called Extended Persistence Landscape.

\begin{definition}[{Extended Persistence Landscape}]
Decompose extended persistence diagram $\text{EDg}$ as $$\text{EDg} = \text{EDg}^{+}\cup \text{EDg}^{-},$$ where the $+$ and $-$ sets contain the points $(b_i,d_i)$ with $b_i< d_i$ and $d_i<b_i$, respectively. Define the $k^{th}$ landscape function $\lambda_k(\mathcal{G})(t)$ as the $k^{th}$ largest value of $\{\Lambda_{i}(t): (b_{i},d_{i})\in {\text{EDg}}^{+} \}$. Similarly, define the $(-j)^{th}$ landscape function $\lambda_{-j}(\mathcal{G})(t)$ as the $j^{th}$ smallest value of $\{\Lambda_{i}(t): (b_{i},d_{i})\in \text{EDg}^{-} \}$. Then the {\it Extended Persistence Landscape} is the set of landscape functions $$\lambda(\mathcal{G})=\{\lambda_k (\mathcal{G})\}\cup \{\lambda_{-j} (\mathcal{G})\}.$$
\end{definition}
%\bedefinitionin{definition}
%Then, the {\it extended persistence landscape} function $\lambda(\Tilde{\mathcal{G}}):[\epsilon_1,\epsilon_\tau]\to \mathbb{R}$ for $t\in [\epsilon_1,\epsilon_\tau]$ is defined as: 
%$$\lambda(\Tilde{\mathcal{G}})(t)=\max_i\Lambda_i(t),$$ 
%where $\{\epsilon_k\}_1^\tau$ are thresholds for the filtration used. 
%\end{definition}
Considering the piecewise linear structure of the function $\lambda_k(\mathcal{G})$, it is completely determined by its values at $2\tau-1$ points, i.e., ${(d_i\pm b_i)}/2\in\{\epsilon_1, \epsilon_{1.5}, \epsilon_2, \epsilon_{2.5}, \dots ,\epsilon_\tau\}$, 
where $\epsilon_{k.5}={(\epsilon_k+\epsilon_{k+1})}/{2}$. Hence, a vector of size $1\times (2\tau-1)$ whose entries the values of this function would suffice to capture all the information needed, i.e.,  $\{b_i,d_i,(b_i+d_i)/2\}$ 

%i.e., $\vec{\lambda}_k=[\lambda_k(\mathcal{G})(\epsilon_1), \lambda_k(\mathcal{G})(\epsilon_{1.5}), \lambda_k(\mathcal{G})(\epsilon_2), \ldots, \lambda_k(\mathcal{G})(\epsilon_\tau)]$. %By changing the type of filtration, we obtain a set of landscapes, i.e., $\Tilde{\textbf{EPL}}^{(p)}_{\mathcal{Q}} = \{\vec{\Tilde{\lambda}}^{(p)}_1, \dots, \vec{\Tilde{\lambda}}^{(p)}_\mathcal{Q}\} \in \mathbb{R}^{\mathcal{Q} \times 1 \times \tau}$.

%\begin{definition}
%Given EPL $\lambda(\Tilde{\mathcal{G}})$, the $p$-norm (where $1\leq p \leq\infty$) in $\lambda(\Tilde{\mathcal{G}})$ is defined by $$\lVert\lambda(\Tilde{\mathcal{G}})\rVert_p = \left(\sum_{k=1}^{\infty}\lVert \lambda_{k}(\Tilde{\mathcal{G}})\rVert_{p}  \right)^{1/p}.$$ \end{definition}
% For $1\leq p \leq\infty$ and $\lambda(\Tilde{\mathcal{G}})$ an extended persistence landscape define {\it the $p$-norm in persistence landscapes} by

\begin{definition}[Distances between EPLs]
    Let ${\text{EDg}}_1$ and ${\text{EDg}}_2$ be two EPDs with corresponding extended persistence landscapes 
    $\lambda({\mathcal{G}}_1)$ and $\lambda({\mathcal{G}}_2)$. For $1\leq p \leq \infty$, the {\it $p$-landscape distance} between ${\text{EDg}}_1$ and ${\text{EDg}}_2$ is defined as
    $$\Lambda_p({\text{EDg}}_1, {\text{EDg}}_2)= \lVert \lambda({\mathcal{G}}_1)-\lambda({\mathcal{G}}_2) \rVert_p,$$ where
    $\lVert \cdot \rVert_p$ is a $\ell_p$-norm.
    Analogously, if ${\text{EM}}_1$ and ${\text{EM}}_2$ are the persistence modules corresponding to ${\text{EDg}}_1$ and ${\text{EDg}}_2$, the {\it $p$-landscape distance} ($1\leq p\leq \infty$) between ${\text{EM}}_1$ and ${\text{EM}}_2$ is defined as
     $$\Lambda_p({\text{EM}}_1, {\text{EM}}_2)= \lVert \lambda(\mathcal{G}_1)-\lambda(\mathcal{G}_2) \rVert_p.$$
\end{definition}

Now consider the case that we have piecewise linear functions $f$ and $g$ inducing filtrations on the simplicial complex $\mathscr{K}$ and taking values in $\mathbb{R}$. Functions $f$ and $g$ define extended persistence modules ${\text{EM}}_1$ and  ${\text{EM}}_2$, extended persistence diagrams ${\text{EDg}}_1$ and  ${\text{EDg}}_2$, as well as extended landscapes $\lambda({\mathcal{G}}_1)$ and $\lambda({\mathcal{G}}_2)$, respectively. Stability of EPD~\cite{cohen2009extending} holds in the following sense: 
\begin{eqnarray}\label{stab}
d_B({\text{EDg}}_1,{\text{EDg}}_2)\leq \lVert f-g \rVert_\infty,
\end{eqnarray}
where $d_{B}$ is the bottleneck distance. In particular, after extending the notion of $\epsilon$-interleaving to the case of extended persistence, it is also true that 
\begin{eqnarray}\label{interleav}
\Lambda_{\infty}({\text{EM}}_1,\text{EM}_2)\leq d_{I}({\text{EM}}_1,\text{EM}_2),
\end{eqnarray}
where $d_{I}$ is the interleaving distance~\cite{Bubenik_2015}. Armed with these results, we state the stability guarantees for EPL.

\begin{proposition}[Stability of EPL] Let ${\text{EDg}}_1$ and ${\text{EDg}}_2$ be EPDs for the piecewise linear functions $f,g:\mathscr{K}\to \mathbb{R}$ respectively, then their corresponding $\infty$-landscape distance satisfies
$$\Lambda_{\infty}({\text{EDg}}_1, {\text{EDg}}_2)\leq d_{B}(\text{EDg}_1, {\text{EDg}}_2)\leq \lVert f-g\rVert_{\infty}.$$
\end{proposition}
Proof of Proposition~0.3 is in Appendix~B.%~\ref{proof}.

{\bf Extended Persistent Image (EPI).} %$\color{gray}{{\text{EDg}}} \rightarrow$ $\color{gray}{{\text{EPI}}}$}: 
Similarly, the extracted EP information can be encoded in a form of the {\it Extended Persistence Image} (EPI), which is an analogue of the ordinary persistence image proposed by ~\cite{adams2017persistence}. EPI has been used in graph learning before, for example, in conjunction with link prediction~\cite{yan2021link}, but has not been formally defined.

EPL is as a finite-dimensional vector representation derived by the weighted kernel density function and can be formulated via the following two steps: \textit{Step 1:} mapping the ${\text{EDg}}$ to an integrable function $\rho_{{\text{EDg}}}: \mathbb{R} \rightarrow \mathbb{R}^{2}$, which is called a persistence surface. The persistence surface $\rho_{{\text{EDg}}}$ is given by sums of weighted Gaussian functions that are centered at each point in ${\text{EDg}}$ and \textit{Step 2:} integrating the persistence surface $\rho_{{\text{EDg}}}$ over each grid box to obtain ${\text{EPI}}$.
Specifically, the value of each pixel $z$ within the ${\text{EPI}}$ is formed as:
\begin{align*}
{\text{EPI}}(z)=\iint\limits_{z} \sum_{\mu \in I} \frac{f(\mu)}{2 \pi \sigma_{x} \sigma_{y}} e^{-\left(\frac{\left(x-\mu_{x}\right)^2}{2\sigma_{x}^{2}}+\frac{\left(y-\mu_{y}\right)^2}{2\sigma_{y}^{2}}\right)} d y d x,
\end{align*}
where $f(\mu)$ is a weighting function (where mean $\mu = (\mu_x, \mu_y) \in \mathbb{R}^2$), and $\sigma_x$ and $\sigma_y$ are the standard deviations in $x$ and $y$ directions. %Lastly, following above steps, we obtain a set of extended persistence images, i.e., $\Tilde{\textbf{EPI}}^{(p)}_{\mathcal{Q}} = \{\Tilde{\text{EPI}}^{(p)}_1, \dots, \Tilde{\text{EPI}}^{(p)}_\mathcal{Q}\} \in \mathbb{R}^{\mathcal{Q} \times \rho \times \rho}$ (where $\rho$ is the resolution size).

\begin{remark}[On theoretical properties of EPI and  relationships to EPL] There are differences in theoretical properties between the two extended persistence summaries presented in the present work, namely, EPL and EPI. As shown by Proposition~4.3, EPLs are stable under the $\infty$-landscape distance and, furthermore, the distance between two EPIs is bounded by the bottleneck distance among the respective EPDs. On the contrary, similar stability result does not hold for EPIs. Indeed, as shown by~\cite{adams2017persistence}, PIs for ordinary persistence are stable only with respect to the 1-Wasserstein distance.
However, at this point there exists a result on stability of EPDs with respect to the bottleneck distance only~\cite{cohen2009extending} and, hence, nothing can be said on stability of EPIs. Nevertheless, stability and
the associated universal distances for extended persistence
is an active research area in algebraic topology~\cite{bauer2022universal}. We then leave this fundamental result as a future fundamental research direction.
\end{remark}

%{\bf Remark} \YGL{Jose and Yuzhou, we introduce two EPH summaries, i.e. EPL and EPI. We need to put something on EPL vs. EPI. In general, EPIs tend to deliver somewhat better performance than EPL. However, EPIs are more computationally costly and do not enjoy the same stability guarantees as EPL.}

{\bf Multiple Filtrations.}
%{$\color{gray}{{\text{Single Filtration}}} \rightarrow$ $\color{gray}{{\text{Multiple Filtrations}}}$}: 
To gain a better understanding of the complex representations of graph data, instead of a single filtration, we hypothesize that learning topological representations via multiple filtrations can benefit neural network framework in gaining both generalization and robustness. Hence, here we consider $\mathcal{Q}$ sublevel filtration functions $\boldsymbol{\mathfrak{{F}}} = \{\mathfrak{F}_1, \dots, \mathfrak{F}_q, \dots, \mathfrak{F}_{\mathcal{Q}}\}$ defined on nodes of ${\mathcal{G}}$ (where $q \in [1, \mathcal{Q}]$ and $\mathcal{Q} \geq 1$). For each filtration $\mathfrak{F}_q$, we obtain an extended persistence diagram denoted by ${\text{EDg}}_q$. Then, by using $\mathcal{Q}$ different filtration functions, we can generate a set of persistence diagrams, i.e., ${\textbf{EDg}}_{\mathcal{Q}} = \{{\text{EDg}}_1, \dots, {\text{EDg}}_\mathcal{Q}\}$. Through the extended persistence vectorization methods above, we can have ${\textbf{EPI}}_{\mathcal{Q}} = \{{\text{EPI}}_1, \dots, {\text{EPI}}_\mathcal{Q}\}$ and ${\textbf{EPL}}_{\mathcal{Q}} = \{{\text{EPL}}_1, \dots, {\text{EPL}}_\mathcal{Q}\}$. For the sake of simplicity, we denote ${\textbf{EPI}}_{\mathcal{Q}}$/${\textbf{EPL}}_{\mathcal{Q}}$ as ${\boldsymbol{\Xi}}$ in the following discussion.

\begin{figure*}[t!]
    \centering%0.44
    \includegraphics[width=1.\textwidth]{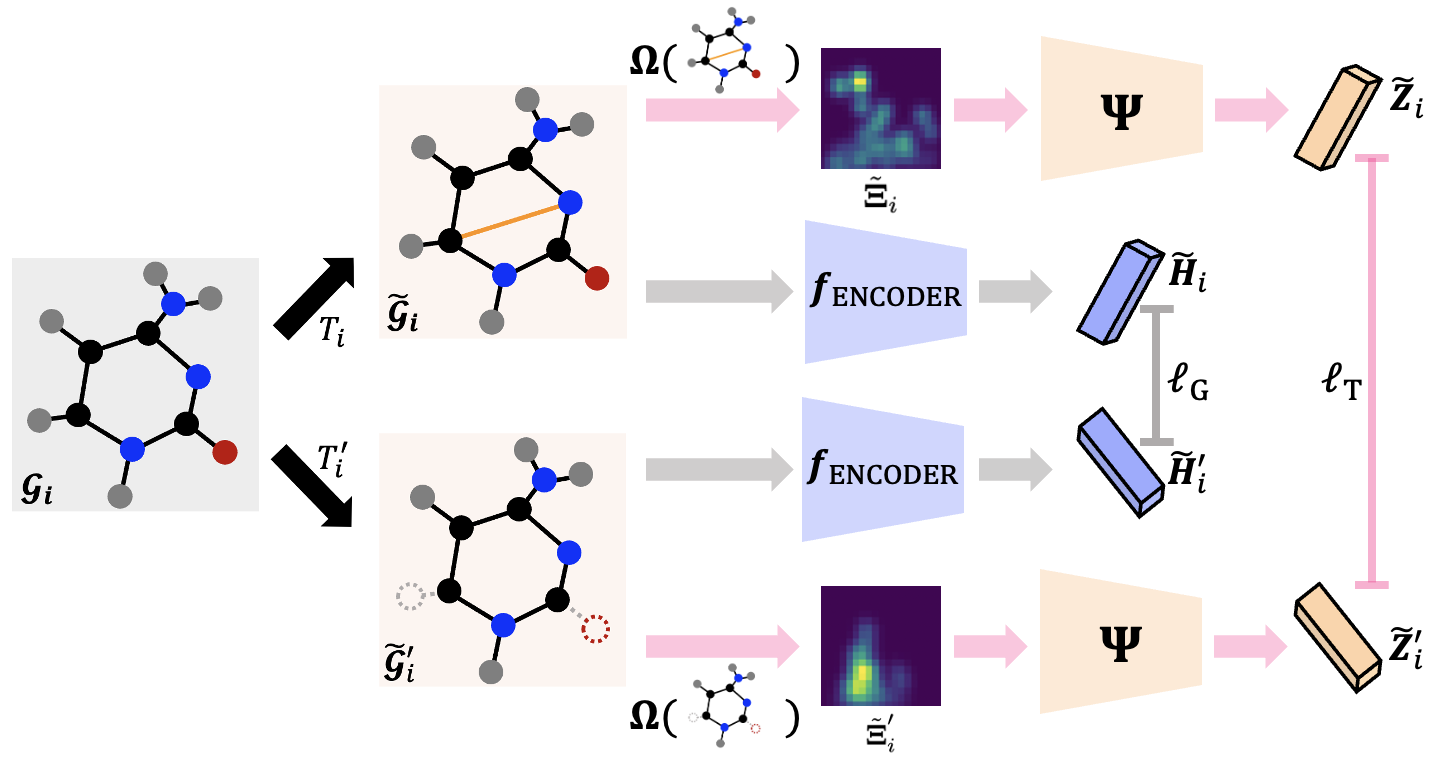}
    %\vspace{-1.5ex}
    \caption{The overall architecture of TopoGCL. TopoGCL consists of 4 components: (I) Calculate an extended topological feature $\Tilde{\boldsymbol{\Xi}}_i$ from the perturbed graph $\mathcal{G}_i$ and then feed $\Tilde{\boldsymbol{\Xi}}_i$ into the the extended topological layer (ETL) $\Psi(\cdot)$ and obtain the latent extended topological representation $\Tilde{\boldsymbol{Z}}_i$. (II) Feed $\mathcal{G}_i$ into the GNN encoder $f_{\text{ENCODER}}$ and generate the node embeddings $\Tilde{\boldsymbol{H}}_i$. (III) Feed $\mathcal{G}^\prime_i$ into the GNN encoder $f_{\text{ENCODER}}$ and generate the node embeddings $\Tilde{\boldsymbol{H}}^\prime_i$. (IV) Calculate an extended topological feature $\Tilde{\boldsymbol{\Xi}}^\prime_i$ from the perturbed graph $\mathcal{G}^\prime_i$ and then feed $\Tilde{\boldsymbol{\Xi}}^\prime_i$ into the the extended topological layer (ETL) $\Psi(\cdot)$ and obtain the latent extended topological representation $\Tilde{\boldsymbol{Z}}^\prime_i$. After that, apply contrastive loss functions (i.e., Equations 1 and 5) to $\{\Tilde{\boldsymbol{H}}_i, \Tilde{\boldsymbol{H}}^\prime_i\}$ and $\{\Tilde{\boldsymbol{Z}}_i, \Tilde{\boldsymbol{Z}}^\prime_i\}$ respectively and obtain two contrastive losses. Finally, combine two contrastive losses via $\ell = \alpha \times \sum^{\Upsilon}_{i=1}\ell_{i, \text{G}} + \beta \times \sum^{\Upsilon}_{i=1}\ell_{i,\text{T}}$.\label{flowchart}}
\end{figure*}

{\bf Extended Topological Layer.} To learn critical information on the extended topological features, we propose the extended topological layer (ETL) denoted by $\Psi$. ETL is a extended topological representation learning layer, illustrated in Figure~\ref{flowchart} (see the {\color{black}{orange}} box). Given extended topological features $\Tilde{\boldsymbol{\Xi}}$, the ETL operator will output the latent extended topological representation $\Tilde{\boldsymbol{Z}}$ with shape $d_c$ as follows ($d_c$ is the number of channels in output):
\begin{align}
\label{extended_topo_layer}
    \Tilde{\boldsymbol{Z}} = \Psi(\Tilde{\boldsymbol{\Xi}}) = \begin{cases}
    \phi_{\text{MAX}}\left(f^{(\ell)}_{\text{CNN}}(\Tilde{\boldsymbol{\Xi}})\right), & \Tilde{\boldsymbol{\Xi}} = \Tilde{\textbf{EPI}}\\
    \text{MLPs}(\Tilde{\boldsymbol{\Xi}}), & \Tilde{\boldsymbol{\Xi}} = \Tilde{\textbf{EPL}}
    \end{cases}, 
\end{align}
where $f^{(\ell)}_{\text{CNN}}$ is the convolutional neural network (CNN) in the $\ell$-th layer, $\phi_{\text{MAX}}$ denotes global max-pooling layer, and MLPs denotes multi-layer perceptrons. Specifically, the ETL provides two types of topological signature embedding functions, i.e., (i) if the input is in the form of an extended persistence image, we use a CNN-based model (e.g., residual networks) to learn the corresponding topological features and employ global max-pooling layer to obtain the image-level feature, and (ii) if the input is the form of extended persistence landscape, we can use MLPs to generate latent topological representation.

{\bf Contrastive Learning on Topological Representations.}
Using the above Equation~\ref{extended_topo_layer}, we generate latent extended topological representations $\Tilde{\boldsymbol{Z}}_i = \Psi(\Tilde{\boldsymbol{\Xi}}_i)$ and $\Tilde{\boldsymbol{Z}}^\prime_i = \Psi(\Tilde{\boldsymbol{\Xi}}^\prime_i)$ from two perturbed graphs $\Tilde{\mathcal{G}}_i$ and $\Tilde{\mathcal{G}}^\prime_i$ (i.e., two augmented views from the same graph $\mathcal{G}$), respectively. Following the previous results of~\cite{you2020graph}, for every latent extended topological representation $\Tilde{\boldsymbol{Z}}_{i}$ being the anchor instance, its positive sample is the latent extended topological representation $\Tilde{\boldsymbol{Z}}^\prime_{i}$ of the another augmented view. Naturally, negative pairs are latent extended topological representations (e.g., $\Tilde{\boldsymbol{Z}}_{\gamma}$) generated from other augmented graphs (e.g., $\Tilde{\mathcal{G}}_{\gamma} \in \Tilde{\boldsymbol{\mathcal{G}}} \setminus \{\Tilde{\mathcal{G}}_i, \Tilde{\mathcal{G}}^\prime_i\}$, where $\Tilde{\mathcal{G}}_{\gamma}$ is an augmented graph of $\mathcal{G}_{\gamma}$ and $\Tilde{\boldsymbol{\mathcal{G}}}$ is a set of $2\Upsilon$ augmented graphs). Then the loss of function is defined to enforce maximizing the consistency between positive pairs $(\Tilde{\boldsymbol{Z}}_{i}, \Tilde{\boldsymbol{Z}}^\prime_{i})$ compared with negative pairs, which is formulated as:
\begin{align}
\label{topo_loss}
    \ell_{i, \text{T}}(\Tilde{\mathcal{G}}_i, \Tilde{\mathcal{G}}^\prime_i) = -\log\frac{\exp{(\text{sim}(\Tilde{\boldsymbol{Z}}_{i}, \Tilde{\boldsymbol{Z}}^\prime_{i})/\zeta)}}{\sum^{2\Upsilon}_{\gamma, \gamma \neq i} \exp{(\text{sim}(\Tilde{\boldsymbol{Z}}_{i}, \Tilde{\boldsymbol{Z}}_{\gamma})/\zeta)}},
\end{align}
where $\text{sim}(\Tilde{\boldsymbol{Z}}_{i}, \Tilde{\boldsymbol{Z}}^\prime_{i}) = \Tilde{\boldsymbol{Z}}^{\top}_{i}\Tilde{\boldsymbol{Z}}^\prime_{i}/||\Tilde{\boldsymbol{Z}}_{i}|| ||\Tilde{\boldsymbol{Z}}^\prime_{i}||$ and $\Tilde{\boldsymbol{Z}}_{\gamma}$ is the latent extended topological representation of $\Tilde{\mathcal{G}}_{\gamma}$ learnt by our proposed ETL. Intuitively, the final training objective function $\ell$ combines Equations~\ref{graph_loss} and~\ref{topo_loss}, i.e., $\ell = \alpha \times \sum^{\Upsilon}_{i=1}\ell_{i, \text{G}} + \beta \times \sum^{\Upsilon}_{i=1}\ell_{i,\text{T}}$, where $\alpha$ and $\beta$ are hyperparameters which balance the contribution of two contrastive losses.

% ${\textbf{Dgs}}_q = \{\text{Dg}^{(0)}_q, \text{Dg}^{(1)}_q, \dots, \text{Dg}^{(p)}_q\}$

%To extract latent topological representation (i.e., from either $\Tilde{\boldsymbol{\Xi}}_i$ or $\Tilde{\boldsymbol{\Xi}}_j$), we propose a topological signature embedding function 

%%%%%%%%%check
\begin{table*}[h!]
\centering
\setlength\tabcolsep{0.5pt}
\footnotesize
%\resizebox{2.1\columnwidth}{!}{
\begin{tabular}{lccccccccc}
\toprule
\textbf{{Model}} &\textbf{{NCI1}} & \textbf{{PROTEINS}} & \textbf{{DD}} & \textbf{{MUTAG}} &\textbf{{DHFR}} & \textbf{BZR} & \textbf{COX2} & \textbf{{PTC\_MR}} & \textbf{{PTC\_FM}}\\
\midrule
GL & N/A & N/A & N/A & 81.66$\pm$2.11 & N/A & N/A & N/A &57.30$\pm$1.40 & N/A\\
WL & 80.01$\pm$0.50 & 72.92$\pm$0.56 & 74.00$\pm$2.20 & 80.72$\pm$3.00 & N/A & N/A & N/A & 58.00$\pm$0.50 & N/A\\
DGK & 80.31$\pm$0.46 & 73.30$\pm$0.82 & N/A & 87.44$\pm$2.72 & N/A & N/A & N/A & 60.10$\pm$2.60 & N/A\\
node2vec & 54.89$\pm$1.61 & 57.49$\pm$3.57 & N/A & 72.63$\pm$10.20& N/A & N/A & N/A & N/A & N/A\\
sub2vec & 52.84$\pm$1.47 & 53.03$\pm$5.55& N/A & 61.05$\pm$15.80& N/A & N/A & N/A & N/A & N/A\\
graph2vec & 73.22$\pm$1.81 & 73.30$\pm$2.05& N/A & 83.15$\pm$9.25& N/A & N/A & N/A & N/A & N/A\\
InfoGraph &76.20$\pm$1.06 & 74.44$\pm$0.31 &72.85$\pm$1.78 & {89.01$\pm$1.13} & \underline{80.48$\pm$1.34} & 84.84$\pm$0.86 & 80.55$\pm$0.51 & 61.70$\pm$1.40 & 61.55$\pm$0.92\\
GraphCL &77.87$\pm$0.41 & 74.39$\pm$0.45 &78.62$\pm$0.40 & 86.80$\pm$1.34 & 68.81$\pm$4.15 & 84.20$\pm$0.86 & \underline{81.10$\pm$0.82} & 61.30$\pm$2.10 & \underline{65.26$\pm$0.59}\\
AD-GCL &73.91$\pm$0.77 & 73.28$\pm$0.46 &75.79$\pm$0.87 & {88.74$\pm$1.85} & 75.66$\pm$0.62 & 85.97$\pm$0.63 & 78.68$\pm$0.56 & \underline{63.20$\pm$2.40} & 64.99$\pm$0.77\\
AutoGCL &{\bf 82.00$\pm$0.29} & {75.80$\pm$0.36} &77.57$\pm$0.60 &88.64$\pm$1.08 & 77.33$\pm$0.76 & \underline{86.27$\pm$0.71} & 79.31$\pm$0.70& {63.10$\pm$2.30} & 63.62$\pm$0.55\\
RGCL &78.14$\pm$1.08 & 75.03$\pm$0.43 & \underline{78.86$\pm$0.48} & 87.66$\pm$1.01 & 76.37$\pm$1.35 & 84.54$\pm$1.67 & 79.31$\pm$0.68 &  61.43$\pm$2.50 & 64.29$\pm$0.32\\
GCL-TAGS & 71.43$\pm$0.49 & \underline{75.78$\pm$0.41} & 75.78$\pm$0.52 & \underline{89.12$\pm$0.76} & N/A & N/A & N/A & N/A & N/A \\
\midrule
\textbf{TopoGCL} &\underline{81.30$\pm$0.27} & $^{***}${\bf 77.30$\pm$0.89} & $^{*}${\bf 79.15$\pm$0.35} & $^{***}${\bf 90.09$\pm$0.93} & $^{***}${\bf 82.12$\pm$0.69} & $^{***}${\bf 87.17$\pm$0.83} & $^{**}${\bf 81.45$\pm$0.55} & {\bf 63.43$\pm$1.13} & $^{***}${\bf 67.11$\pm$1.08}\\
\bottomrule
\end{tabular}%}
%\vspace{-1.5ex}
\caption{Performance on molecular and chemical graphs.\label{classification_results_bio_graphs}}
\end{table*}
%%%%%%%%%

\begin{table}[h!]
\centering
\footnotesize
%\resizebox{1.\columnwidth}{!}{
\begin{tabular}{lccc}
\toprule
\textbf{{Model}} &\textbf{{IMDB-B}} & \textbf{{IMDB-M}} & \textbf{{REDDIT-B}} \\
\midrule
GL & 65.87$\pm$0.98 & 46.50$\pm$0.30 & 77.34$\pm$0.18\\
WL & 72.30$\pm$3.44 & 47.00$\pm$0.50 & 68.82$\pm$0.41\\
DGK & 66.96$\pm$0.56 & 44.60$\pm$0.50 & 78.04$\pm$0.39\\
%MLG &\\
%\midrule
node2vec &  56.40$\pm$2.80 & 36.00$\pm$0.70 & 69.70$\pm$4.10 \\
sub2vec & 55.26$\pm$1.54 & 36.70$\pm$0.80 & 71.48$\pm$0.41 \\
graph2vec & 71.10$\pm$0.54 & 50.40$\pm$0.90 & 75.78$\pm$1.03 \\
InfoGraph & {73.03$\pm$0.87} & 49.70$\pm$0.50 & 82.50$\pm$1.42\\
GraphCL & 71.14$\pm$0.44 & 49.20$\pm$0.60 &89.53$\pm$0.84\\
AD-GCL & 70.21$\pm$0.68 & {50.60$\pm$0.70} & 90.07$\pm$0.85\\
AutoGCL & 72.32$\pm$0.93 & 50.60$\pm$0.80 &88.58$\pm$1.49 \\
RGCL & 71.85$\pm$0.84 & 49.31$\pm$0.42 & \underline{90.34$\pm$0.58}\\
GCL-TAGS &  \underline{73.65$\pm$0.69} & \underline{52.16$\pm$0.72} & 83.62$\pm$0.64\\
\midrule
\textbf{TopoGCL} & $^{***}${\bf 74.67$\pm$0.32} & {\bf 52.81$\pm$0.31} & {\bf 90.40$\pm$0.53} \\
\bottomrule
\end{tabular}%}
%\vspace{-1.5ex}
\caption{Performance on social graphs.\label{classification_results_social_graphs}}
\end{table}

\section{Experiments}
{\bf Datasets and Baselines.} We validate TopoGCL on unsupervised representation learning tasks using the following 12 real-world graph datasets: (i) 5 chemical compound datasets: NCI1, MUTAG, DHFR, BZR, and COX2, (ii) 4 molecular compound datasets: DD, PROTEINS, PTC\_MR, and PTC\_FM, (iii) 2 internet movie databases: IMDB-BINARY (IMDB-B) and IMDB-MULTI (IMDB-M), and (iv) 1 Reddit discussion threads dataset: REDDIT-BINARY (REDDIT-B). Each dataset includes multiple graphs of each class, and we aim to classify graph classes. The statistics of the 12 graph datasets are shown in Appendix~D.1, Table~\ref{datasets}. For all graphs, following the experimental settings of GraphCL~\cite{you2020graph}, we use 10-fold cross validation accuracy as the classification performance (based on a non-linear SVM model, i.e., LIB-SVM~\cite{chang2011libsvm}) and repeat the experiments 5 times to report the mean and standard deviation. The best results are given in {\bf bold} while the best performances achieved by the runner-ups are \underline{underlined}. We also conduct a one-sided two-sample $t$-test between the best result and the best performance achieved by the runner-up, where *, **, *** are $p$-value $<$ 0.1, 0.05, 0.01, i.e., significant, statistically significant, highly statistically significant results, respectively. We evaluate the performances of our TopoGCL on 12 graph datasets versus 12 state-of-the-art baselines including: (i) Graphlet Kernel (GL)~\cite{shervashidze2009efficient},  (ii) Weisfeiler-Lehman Sub-tree Kernel (WL)~\cite{shervashidze2011weisfeiler}, (iii) Deep Graph Kernels (DGK)~\cite{yanardag2015deep}, (iv) node2vec~\cite{grover2016node2vec}, (v) sub2vec~\cite{adhikari2018sub2vec}, (vi) graph2vec~\cite{narayanan2017graph2vec}, (vii) InfoGraph~\cite{sun2019infograph}, (viii) GraphCL~\cite{you2020graph}, (ix) AD-GCL~\cite{suresh2021adversarial}, (x) AutoGCL~\cite{yin2022autogcl}, (xi) RGCL~\cite{li2022let}, and (xii) GCL-TAGS~\cite{lin2022spectrum}. 

{\bf Experiment Settings.}
We conduct our experiments on two NVIDIA RTX A5000 GPU cards with 24GB memory. TopoGCL is trained end-to-end by using Adam optimizer. The tuning of TopoGCL on each dataset is done via grid hyperparameter configuration search over a fixed set of choices and the same cross-validation setup is used to tune baselines. Table~\ref{running_time} in Appendix~D.1 shows the average running time of extended persistence image (EPI) and extended persistence landscape (EPL) generation (in seconds) on all 12 graph datasets. See Appendix~D for more details. The source code of TopoGCL is publicly available at~\url{https://github.com/topogclaaai24/TopoGCL.git}. See Appendix~D for more details.%We consider sublevel filtration functions based on 4 centrality measures for nodes including degree, betweenness, closeness, and subgraph centralities, and search the optimal number of sublevel filtration functions $\mathcal{Q} \in \{1,2,3,4\}$. In our experiments, we select the filtration functions via the cross-validation. In all experiments, we use the CNN-based model and MLPs to learn EPIs and EPLs respectively. 

\subsection{Experiment Results}
The evaluation results on 12 graph datasets are summarized in Tables~\ref{classification_results_bio_graphs} and~\ref{classification_results_social_graphs}. We also conduct ablation studies and robustness analysis to assess the contributions of the extended persistence and the robustness of TopoGCL against noisy scenarios.
%adversarial attacks.

{\bf Molecular, Chemical, and Social Graphs.} Table~\ref{classification_results_bio_graphs} shows the performance comparison among 12 baselines on NCI1, PROTEINS, DD, MUTAG, DHFR, BZR, COX2, and two PTC datasets with different carcinogenicities on rodents (i.e., PTC\_MR, and PTC\_FM) for graph classification. Our TopoGCL is comparable with the state-of-the-art on the NCI1 dataset, and consistently outperforms baseline models on other 8 datasets. In particular, the average relative gain of TopoGCL over the runner-ups is 1.25\%. The results demonstrate the effectiveness of TopoGCL. In terms of unsupervised graph-level representation learning baselines, e.g., node2vec only focuses on the graph structure learning and generates node embeddings through random walks. In turn, InfoGraph works by learning graph representation via GNNs, and taking graph representation and patch representation as pairs for unsupervised representation learning,  hence, resulting in improvement over random walk-based graph embedding methods. Comparing with InfoGraph, graph contrastive learning methods such as GraphCL and RGCL explore the view augmentations approaches and learn representations of augmented graph structures for graph contrastive learning. A common limitation of these approaches is that they do not simultaneously capture both topological properties of the graph and information from the graph structure. Hence, it is not surprising that performance of our proposed TopoGCL which systematically integrates all types of the above information on the observed graphs is substantially higher than that of the benchmark models. Besides, we have conducted a visual experiment and the corresponding validity evaluation of the extracted topological and geometric information and its role in GCL (see Appendix~D for a discussion). Table~\ref{classification_results_social_graphs} shows the performance comparison on 3 social graph datasets. Similarly, Table~\ref{classification_results_social_graphs} indicates that our TopoGCL model is always better than baselines for all social graph datasets. Moreover, we find that TopoGCL also has the smallest standard deviation across 3 social graph datasets, revealing that local topological representation learning module can enhance the model stability.

{\bf Ablation Study of the Topological Signatures.}
We perform ablation studies on MUTAG, PTC\_MR, and IMDB-B to justify the following opinions: (i) the benefit of the extended topological signatures in topological representation learning; and (ii) measuring the similarity of positive samples and the diversity between negative samples via both {global-global CL} (i.e., GCL) and {topo-topo CL} can achieve better performance than only considering {Global-Global CL}. For opinion (i), we compare TopoGCL (i.e., TopoGCL + EPI/EPL) with its variant (i.e, TopoGCL + PI/PL). Note that, for PTC\_MR, we use EPL/PL due to TopoGCL based on persistence landscapes achieves better performance than persistence images. Results are shown in Table~\ref{ablation_study_1}, Appendix~E. We observe that TopoGCL based on extended topological signatures performs much better than the variant, demonstrating that our extended topological signatures can capture higher-order structural and topological information efficiently. Note that we found TopoGCL can achieve more competitive results on PTC\_MR by using EPL instead of EPI; thus we specially compare TopoGCL+EPL with TopoGCL+PL in this ablation study. Besides, we validate TopoGCL by comparing TopoGCL's performance with the performance of TopoGCL W/o Topo and the experimental results in Table~\ref{ablation_study_2} in Appendix~E show that topo-topo CL can bring performance gain since the enhancement of structural and topological information learning. 

\begin{table}
%\begin{wraptable}[10]{r}{0.48\textwidth}
%\vspace*{-2pt} % keep vspace here
\centering
\footnotesize % font smaller
%\tabcolsep{1.5pt}
%\renewcommand{\arraystretch}{0.2}
\begin{tabular}{lccc}
\toprule
& \textbf{\footnotesize AD-GCN} & \textbf{\footnotesize RGCL} & \textbf{\footnotesize TopoGCL (ours)} \\
\midrule
\multirow{1}{*}{\textbf{\footnotesize MUTAG}}& 88.20$\pm$1.24 & 87.05$\pm$1.16 & $^{***}${\bf 89.47$\pm$1.04}\\
\midrule
\multirow{1}{*}{\textbf{\footnotesize PTC\_MR}}& 60.88$\pm$0.92 &61.10$\pm$1.47 & {\bf 61.63$\pm$1.09}\\
\midrule
\multirow{1}{*}{\textbf{\footnotesize IMDB-B}}&70.01$\pm$0.86 &70.59$\pm$0.34 & $^{***}${\bf 72.18$\pm$0.15}\\
\bottomrule
\end{tabular}
%\vspace{-1.5ex}
\caption{Robustness study against noise.\label{robustness_study}}
%\vspace{-500pt}
%\vspace*{-20mm} 
\end{table}

{\bf Robustness Study.}
To evaluate the robustness of our proposed TopoGCL under noisy conditions, in this section, we consider adding Gaussian noise into node features of 20\% data. Note that the added noise follows i.i.d Gaussian density, i.e., $\mathcal{N}(1,1)$ (where $\mu = \sigma = 1$). The performances of TopoGCL and other baselines (i.e., AD-GCL and RGCL) on MUTAG, PTC\_MR, and IMDB-B are shown in Table~\ref{robustness_study}. We observe that our TopoGCL always outperforms two state-of-the-art baselines on all three datasets. The strong performance verifies the superiority of TopoGCL and demonstrates that TopoGCL  can efficiently exploit the underlying topological features of the input graph structure.

\section{Conclusion}
We have proposed a novel contrastive learning model, i.e., Topological Graph Contrastive Learning (TopoGCL). TopoGCL adopts a new contrasting mode (topo-topo CL) which can capture both local and global latent topological information. Our extensive experimental results have demonstrated that TopoGCL achieves impressive improvements over the state-of-the-art GCL models in terms of accuracy and enhances the robustness against noisy scenarios. In the future, we will advance 
TopoGCL and the ideas of extended persistence to self-supervised learning of time-evolving graphs, with a particular focus on streaming scenarios. 

\section*{Acknowledgements}
This work was supported by the NSF grants  ECCS 2039701, TIP-2333703, and ONR grant 
N00014-21-1-2530. Also, the paper is based upon work supported by (while Y.R.G. was serving at) the NSF. The views expressed in the article do not necessarily represent the views of NSF or ONR.

\bibliography{AAAI_2024/topogcl}

%\clearpage
\appendix
\section{A. Notation and Details of TopoGCL Architecture}
\subsection{A.1 Notation}
Frequently used notation is summarized in Table~\ref{notations}.

\begin{table*}[ht!]
\centering
\setlength\tabcolsep{1pt}
\begin{tabular}{ll}
\toprule
\textbf{Notation} &  \textbf{Definition} \\
\hline
$\mathcal{G}$ & an attribute graph\\
$\mathcal{V}$ & a set of nodes\\
$\mathcal{E}$ & a set of edges\\
$\boldsymbol{X}$ & a node feature matrix\\
$\boldsymbol{A}$ & an adjacency matrix\\
$\boldsymbol{D}$ & a degree matrix corresponding to $\boldsymbol{A}$\\
$N$ & the number of nodes\\
$F$ & the dimension of node features\\
$\Upsilon$ & number of graphs in a set of graphs \\
$b_\rho$ and $d_\rho$ & a birth time and death time for a topological feature $\rho$\\
$\mathscr{K}$ & an abstract simplicial complex\\
$\mathcal{G}_{\nu_j}$ &  a subgraph with a scale parameter $\nu_j$ in a sequence of nested
subgraphs\\
$\omega$ & a edge-weight function\\
$\mathcal{T}(\cdot)$/$\mathcal{T}_i(\cdot)$/$\mathcal{T}^\prime_i(\cdot)$ & graph data augmentations\\
$f_{\text{ENCODER}}$ & a shared encoder for graph representation learning\\
$\text{sim}(\cdot, \cdot)$ & a similarity function\\
$\boldsymbol{\Tilde{H}}_i$ and $\boldsymbol{\Tilde{H}}^\prime_i$ & learned representations of the two augmented graph $\mathcal{\Tilde{G}}_i$ and $\mathcal{\Tilde{G}}^\prime_i$\\
$\boldsymbol{\Tilde{Z}}_i$ and $\boldsymbol{\Tilde{Z}}^\prime_i$ & latent extended topological representations of the two augmented graph $\mathcal{\Tilde{G}}_i$ and $\mathcal{\Tilde{G}}^\prime_i$\\
$\mathcal{Q}$ & number of sublevel filtration functions\\
$\mathfrak{F}_q$ & the $q$-th sublevel filtration\\
$\Psi(\cdot)$ 
& a extended topological layer (ETL)\\
$\boldsymbol{\Tilde{\Xi}}$ & extended topological features based on a augmented graph $\mathcal{\Tilde{G}}$\\
$\boldsymbol{\Omega}(\cdot)$ & a function which extracts extended persistence features\\
$f_{\text{CNN}}$ & the convolutional neural network\\
$\phi_{\text{MAX}}$ & the global max-pooling layer\\
$\ell_{i, \text{G}}$ & graph contrastive loss for a graph $\mathcal{G}_i$ (in Figure~1 of the main body, we use $\ell_{\text{G}}$ denotes $\ell_{i, \text{G}}$)\\
$\ell_{i, \text{T}}$ & topological contrastive loss for a graph $\mathcal{G}_i$ (in Figure~1 of the main body, we use $\ell_{\text{T}}$ denotes $\ell_{i, \text{T}}$)\\
$\ell$ & the final training objective function\\
$\zeta$ & a temperature hyperparameter\\
$\alpha$ and $\beta$ & hyperparameters which balance the contribution of graph and topological contrastive losses\\
$\mathscr{K}(\mathcal{G}_{\nu_j})$ & the simplicial complex associated to the graph $\mathcal{G}_{\nu_j}$\\

$\Lambda_{i}$ & generating function of extended persistence homology\\
 $\lambda_{k}(\mathcal{G})$ & $k^{th}$ landscape fuction of graph $\mathcal{G}$\\
 $\Lambda_p(\text{EDg}_1, \text{EDg}_2)$  & $\ell_p$-norm between extended persistence diagrams $\text{EDg}_1$ and $\text{EDg}_2$\\
 $\Lambda_p(\text{EM}_1, \text{EM}_2)$  & $\ell_p$-norm between extended persistence modules $\text{EM}_1$ and $\text{EM}_2$\\

 $d_B({\text{EDg}}_1,{\text{EDg}}_2)$ &  bottleneck distance  between extended persistence diagrams $\text{EDg}_1$ and $\text{EDg}_2$\\
 $ d_{I}({\text{EM}}_1,\text{EM}_2)$ & interleaving distance  between extended persistence modules $\text{EM}_1$ and $\text{EM}_2$\\
\midrule
EDg& extended persistence diagram\\
EPI& extended persistence image\\
EPL& extended persistence landscape\\
EM & persistence module\\
%ETL & extended topological layer\\
%$F_N$ & the feature dimension \\
\bottomrule
\end{tabular}
\caption{The main symbols and definitions in this paper.\label{notations}}
\end{table*}

\subsection{A.2 Additional Details of TopoGCL Architecture}

The overall architecture of TopoGCL contains 3 parts (see Figure~\ref{flowchart}): (i) Given a original graph $G_i$, we first apply two graph data augmentations $T_i$ and $T^\prime_i$ on $G_i$ and obtain 2 graph views $\Tilde{G}_i$ and $\Tilde{G}^\prime_i$. (ii) We feed $\Tilde{G}_i$ and $\Tilde{G}^\prime_i$ into both graph and topological contrastive learning channels; more specifically, for graph contrastive learning, we feed $\Tilde{G}_i$ and $\Tilde{G}^\prime_i$ into the shared encoder $f_\text{ENCODER}$ to extract the graph representations $\Tilde{H}_i$ and $\Tilde{H}^\prime_i$; for topological contrastive learning, we first use our extended persistence method (denoted as $\Omega(\cdot)$) to extract extended topological features $\Tilde{\Xi}_i$ and $\Tilde{\Xi}^\prime_i$ from $\Tilde{G}_i$ and $\Tilde{G}^\prime_i$ respectively; then we feed $\Tilde{\Xi}_i$ and $\Tilde{\Xi}^\prime_i$ into the extended topological layer (ETL) $\Psi(\cdot)$ (see Equation 4 in the main body) and extract the topological representations $\Tilde{Z}_i$ and $\Tilde{Z}^\prime_i$. (iii) We then use Equation~1 as contrastive loss functions to enforce maximizing the consistency between positive pairs between $\Tilde{H}_i$ and $\Tilde{H}^\prime_i$, and use Equation~5 as contrastive loss functions to enforce maximizing the consistency between positive pairs between $\Tilde{Z}_i$ and $\Tilde{Z}^\prime_i$. Note that, the final training objective function combines Equations 1 and 5.

\section{B. Proof of Proposition~4.3}
\label{Proposition_proof_43}
\begin{proof}
We partition extended persistent diagram ${\text{EDg}}_1$ as
$${\text{EDg}}_1=Ord(f)\cup Ext(f)\cup Rel(f).$$ That is, we decompose ${\text{EDg}}_1$ into its ordinary, extended and relative subdiagrams. We can also further decompose as $$Ext(f)=Ext(f)^+\cup Ext(f)^-,$$ where $Ext(f)^+$ contains the points $(b_{i},d_{i})$ with $b_{i}<d_{i}$ and $Ext(f)^-$  contains those points with $d_{i}<b_{i}$. We then express $\text{EM}_1$ as the direct sum of 
\begin{eqnarray*}
&&\oplus_{(a,b)\in Ord(f)}\mathbb{I}(a,b), 
\oplus_{(c,d)\in Rel(f)}\mathbb{I}(d,c), \\
&&\oplus_{(x,y)\in Ext^+(f)}\mathbb{I}(x,y), \text{ and } 
\oplus_{(z,w)\in Ext^-(f)}\mathbb{I}(w,z).
\end{eqnarray*}
%\bigoplus_{(a,b)\in Ord(f)}\mathbb{I}(a,b)$, $\bigoplus_{(c,d)\in Rel(f)}\mathbb{I}(d,c)$,  $\bigoplus_{(x,y)\in Ext^+(f)}\mathbb{I}(x,y)$ and $\bigoplus_{(z,w)\in Ext^-(f)}\mathbb{I}(w,z)$. 
We also have a similar decomposition for $\text{EDg}_2$. 

Then, from Theorem 4.9 in~\citet{chazal2016structure}, we have 
\begin{eqnarray}
\label{stab_mod}
d_{I}(\text{EM}_1,\text{EM}_2)\leq d_{B}(\text{EDg}_1,\text{EDg}_2).
\end{eqnarray}
The resulting stability of EPL then follows from combining inequalities~2, ~3 (see inequalities 2 and 3 in the main body) and ~(\ref{stab_mod}).  
%\YGL{Jose, refer to the theorem ref?}.
\end{proof}

\section{C. Background on Extended Persistent Homology}
%\YC{I will put previous rebuttal things on it. We have mentioned that we will provide in the main body.}

\begin{figure*}%{r}{1.\textwidth}
\vspace{-10pt}
\begin{center}
\includegraphics[width=1.\textwidth]{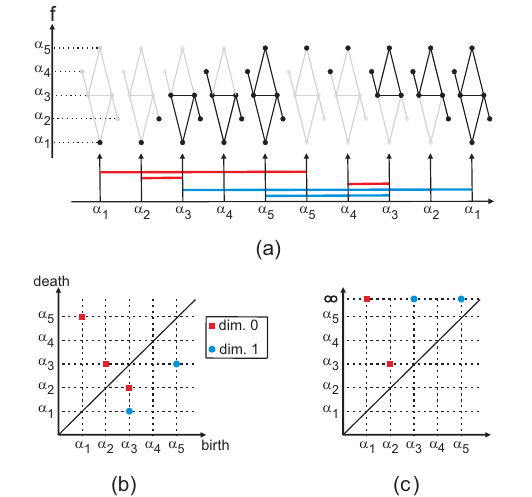}
\end{center}
%\vspace{-20pt}
\caption{Comparison between traditional persistence and extended persistence on a graph.\label{fig:EstendedPers}}
\vspace{-10pt}
\end{figure*}
Persistent homology (PH) is a powerful tool to extract topological and geometric structures in the observed data at various resolution scales. However, the ordinary PH has a number of shortcomings in detecting the relevant information as we will show below, and extended persistent homology (EPH) addresses some of these limitations. We present an illustrative example contrasting PH and EPH in Figure~\ref{fig:EstendedPers}. 
%presents an illustrative example with the general construction of extended persistence diagrams (EPD) in graphs, and we contrast the information encoded in EPDs with that of the classical persistence diagrams (PDs). 
In particular, suppose that a real-valued continuous  function $f$ is defined on a graph $\mathcal{G}$ (the height function in Figure \ref{fig:EstendedPers} a). For a real-value $\alpha$, the sets $f^{-1}(-\infty,\alpha]$ and $f^{-1}[\alpha,+\infty)$ are called the sublevel and superlevel sets of $f$ at $\alpha$. After increasing the value $\alpha$, a filtration on $\mathcal{G}$ is produced. PH detects topological features in different dimensions: connected components (dimension 0)  and loops (dimension 1). The evolution of these topological features is summarized in a persistence diagram (PD) (see Figure \ref{fig:EstendedPers}c), in which every point with coordinates $(b,d)$ represents the filtration value at which a topological feature appears ($b$) and vanishes ($d$). Note that in our example there are three topological features that {\it never} vanish (i.e., the two loops in dimension 1 and the 0-dimensional one). 

However, in EPH, after the filtration obtained from increasing $\alpha$, a filtration given by superlevel sets (decreasing $\alpha$) should be undertaken (to be more precise, this filtration should be considered in relative homology). These two filtrations are schematically represented in Figure \ref{fig:EstendedPers}a. Note that the topological features in PH (Figure \ref{fig:EstendedPers}c) vanish in EPH after going through the downwards filtration (see the second part of diagram in Figure \ref{fig:EstendedPers}a). As a result, the lifespans for EPH (i.e., an absolute difference between birth and death of a topological feature) is finite. This property is very important since the lifespans typically serve as an input to a topological layer of DL models. Clearly, infinite lifespans cannot be input into a DL and arbitrary truncation leads to an information loss. EPH bypasses this problem. Furthermore, in PH, only the branch pointing downwards in the graph (see Figure~\ref{fig:EstendedPers}) is detected. However, in EPH, both branches of the graph are detected (i.e., the two points in dimension 0 with shorter lifespan). 

In summary, the two important advantages of EPH over PH  that were observed in the example are: 
\begin{itemize}
\item[(i)] All topological features have a finite lifespan in EPH but not in ordinary PH. Hence, EPH is more suitable as input to a trainable topological layer in DL. 
\item[(ii)] EPH extracts more topological information than PH and avoids loss of potentially valuable knowledge on the graph. 
\end{itemize}

\begin{figure*}%{r}%{1.\textwidth}
\vspace{-10pt}
\begin{center}
\includegraphics[width=1.\textwidth]{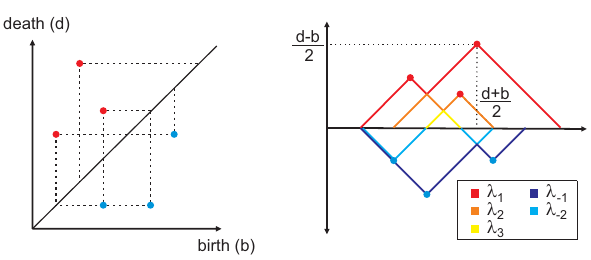}
\end{center}
%\vspace{-20pt}
\caption{Extended persistence landscape.\label{fig:ExtendedLascape}}
\vspace{-10pt}
\end{figure*}

%\begin{wrapfigure}{r}{0.5\textwidth}
%\vspace{-10pt}
\begin{figure*}[h!]
\begin{center}
\includegraphics[width=1.\textwidth]{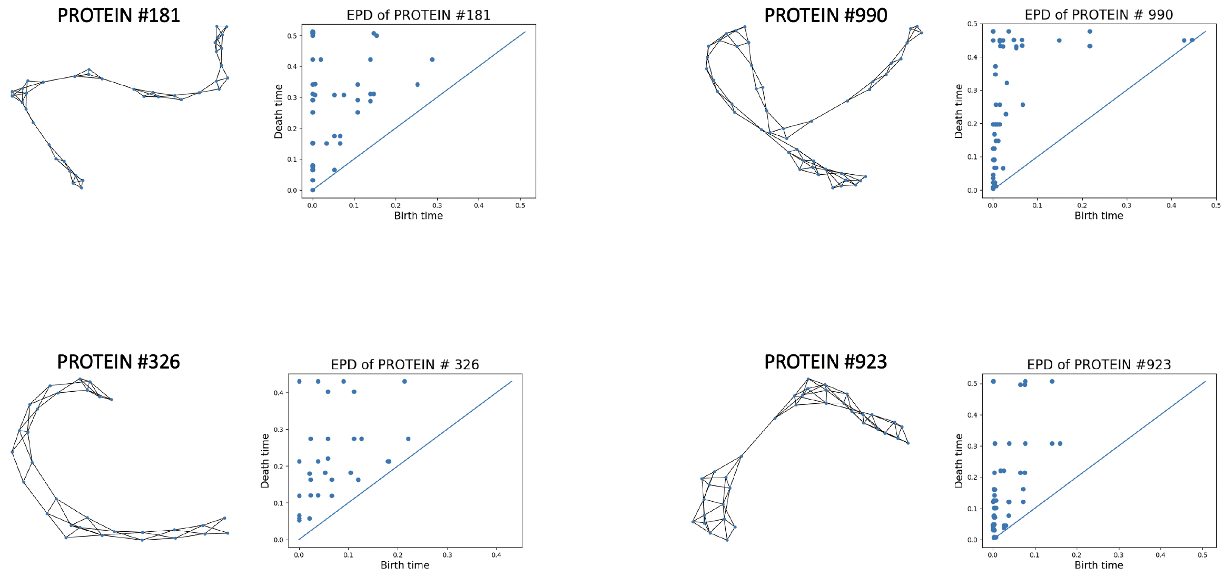}
\end{center}
%\vspace{-20pt}
\caption{Proteins structures and EPDs in PROTEINS.\label{epd_proteins_plot}}
\end{figure*}
%\vspace{-10pt}
%\end{wrapfigure}

We can go further to the definition of Extended Persistence Landscapes (see Figure~\ref{fig:ExtendedLascape} for an illustrative example).  In ordinary PH, all points in a persistence diagram are above the diagonal as  can be seen in Figure \ref{fig:EstendedPers}c. In the case of EPH, this is not always the case as you can see in Figure~\ref{fig:EstendedPers}b. In order to extend the classical definition of persistence diagram to the case EPH, we distinguish between two types of landscape functions, those that are above the $x$ axis and which are bellow as in Figure \ref{fig:ExtendedLascape}. 

%\onecolumn
%\section{You \emph{can} have an appendix here.}
%You can have as much text here as you want. The main body must be at most $8$ pages long. For the final version, one more page can be added. If you want, you can use an appendix like this one, even using the one-column format.
%\section{Datasets}

\section{D. Datasets, Experiment Settings, and Additional Experiments}

In this paper, we conduct extensive comparisons with state-of-the-art baselines for graph classification tasks. Note that, the same idea and methodology can be extended to both node classification and link prediction tasks~\cite{wu2021self,liu2022graph}, but these future directions go beyond the scope of a single paper.

\subsection{D.1. Datasets and Experiment Settings}
\label{dataset_exper_settings}
We validate TopoGCL on unsupervised representation learning tasks using the following 12 real-world graph datasets: (i) 5 chemical compound datasets: NCI1, MUTAG, DHFR, BZR, and COX2; (ii) 4 molecular compound datasets: DD, PROTEINS, PTC\_MR, and PTC\_FM; (iii) 2 internet movie databases: IMDB-BINARY (IMDB-B) and IMDB-MULTI (IMDB-M); and (iv) 1 Reddit (an online aggregation and discussion website) discussion threads dataset: REDDIT-BINARY (REDDIT-B).
\begin{table}[h!]
\centering
%\resizebox{0.8\columnwidth}{!}{
%\setlength\tabcolsep{4pt}
\begin{tabular}{lccccc}
\toprule
\textbf{{Dataset}} & \textbf{{\# Graphs}} &\textbf{{Avg.} $|\mathcal{V}|$} & \textbf{{Avg.} $|\mathcal{E}|$} & \textbf{{\# Class}} \\
\midrule
NCI1 & 4110 & 29.87 & 32.30 & 2\\
PROTEINS &1113 &39.06 &72.82 &2 \\
DD & 1178 & 284.32 & 715.66 & 2\\
MUTAG &188 &17.93 &19.79 &2 \\
DHFR & 467 & 42.43 & 44.54 & 2\\
BZR &405 &35.75 &38.35 &2 \\
COX2 &467 &41.22 &43.45 &2 \\
PTC\_MR &  344 & 14.29 & 14.69 & 2\\
%PTC\_MM &  336 & 13.97 & 14.32 & 2\\
PTC\_FM &  349 & 14.11 & 14.48 & 2\\
%PTC\_FR & 351 & 14.56 & 15.00 & 2 \\
IMDB-B &1000 &19.77 &96.53 &2 \\
IMDB-M & 1500 & 13.00 & 65.94 & 3 \\
REDDIT-B & 2000 &  429.63 & 497.75& 2 \\
\bottomrule
\end{tabular}%}
\caption{Summary statistics of the benchmark datasets.\label{datasets}}
\end{table}

Table~\ref{running_time} reports the average running time (seconds) of each extended persistence image (EPI) and extended persistence landscape (EPL) generation on all 12 graph datasets.
We find that
EPL tends to be somewhat more computationally efficient than EPI, while their classification performances are comparable. This makes EPL a competitive summary for larger scale datasets.

\begin{table}[t!] %[b]
\centering
\begin{tabular}{lcc}
\toprule
\multirow{2}{*}{\textbf{Dataset}} & \multicolumn{2}{c}{\textbf{Average Time Taken (sec)}} \\
& EPI & EPL \\
\midrule
NCI1 & 0.0757 & 0.0681\\
PROTEINS & 0.0668 & 0.0586\\
DD & 1.3801 & 1.2640\\
MUTAG & 0.0219 & 0.0201\\
DHFR & 0.0439 & 0.0352\\
BZR & 0.0390 & 0.0310\\
COX2 & 0.0398 & 0.0332\\
PTC\_MR &0.0216  & 0.0185 \\
PTC\_FM & 0.0229 & 0.0179\\
IMDB-B & 0.0349 & 0.0333\\
IMDB-M & 0.0220 & 0.0196\\
REDDIT-B & 1.0712 & 0.9765\\
\bottomrule
\end{tabular}
\caption{Computational costs for generation of the extended persistence image (EPI) and extended persistence landscape (EPL).\label{running_time}}
\end{table}

We conduct our experiments on two NVIDIA RTX A5000 GPU cards with 24GB memory. TopoGCL is trained end-to-end by using Adam optimizer. The tuning of TopoGCL on each dataset is done via grid hyperparameter configuration search over a fixed set of choices and the same cross-validation setup is used to tune baselines. In our experiments, for all datasets, (i) $\Tilde{\text{EPI}}$: we set the grid size of $\Tilde{\text{EPI}}$ to $50 \times 50$ (i.e., $\rho=50$), and (ii) $\Tilde{\text{EPL}}$: we set the number of piecewise linear functions to output and number of sample for all piecewise-linear functions to 2 and 50 respectively. We utilize `extended\_persistence()' function in GUDHI to extract extended persistence diagrams (EPDs). Regarding the extended persistence image (EPI), we follows steps 1-2 in Extended Persistent Image (EPI) part (see lines \#235-\#238 in the main body) to build the pipeline; for the extended persistence landscape (EPL), we use the `Landscape' function in GUDHI (from gudhi.representations). We consider sublevel filtration functions based on 4 centrality measures for nodes including degree, betweenness, closeness, and subgraph centralities, and search the optimal number of sublevel filtration functions $\mathcal{Q} \in \{1,2,3,4\}$. To extract valuable topological features for representation learning, we consider using multifilrations, i.e., combining topological features computed from multiple different filtration functions (which can capture certain topological features from different perspectives). Moreover, we search for the optimal filtration functions combination through cross-validation. From experiments, we find that node betweenness- and node closeness-based filtration functions always play important roles. More specifically, to evaluate the importance of different filtration functions, we apply an attention mechanism on topological features of multifiltrations and find the importance weights of topological features from node betweenness- and node closeness-based filtration functions are higher than others. In all experiments, we use the CNN-based model and MLPs to learn extended persistence images and extended persistence landscapes respectively. More specifically, the CNN-based model consists
of 2 CNN layers with number of hidden unit to 32, kernel size to 2, stride to 2, and the global max-pooling with the pool size of $5 \times 5$; MLPs backbone consists of 5 layers. We use node drop augmentation for graph datasets with a drop ratio of 0.1. In contrastive loss function, we set the temperature hyperparameter $\zeta$ to be 0.2. Moreover, the backbone of $f_{\text{ENCODER}}(\cdot)$ consists of a graph isomorphism networks (GINs)~\cite{xu2018powerful} and a projection head. In our experiments, we perform an extensive grid search for hyperparameters $\alpha$ and $\beta$ over the search space $\alpha, \beta \in \{0.1, 0.2, \dots, 1.0\}$.

%%%%%%%%%%%%%%%%%%%%%%%%%%%%%%%%%%%%%%%%%%%%%%%%%%%%%%%%%%%%%%%%%%%%%%%%%%%%%%%
%%%%%%%%%%%%%%%%%%%%%%%%%%%%%%%%%%%%%%%%%%%%%%%%%%%%%%%%%%%%%%%%%%%%%%%%%%%%%%%
\subsection{D.2. Additional Experiments}\label{appendix_d_2}
We have conducted a visual experiment and the corresponding validity evaluation of the extracted topological and geometric information and its role in GCL. Figure~\ref{epd_proteins_plot} shows 4 protein networks (i.e., Protein\_181, Protein\_326, Protein\_923, Protein\_990 from the PROTEINS dataset), along with their corresponding extended persistence diagrams (EPDs). We find that it appears quite hard to identify which protein networks belong to the same class (i.e., enzyme or non-enzyme) from the conventional graph structure characteristics. For instance, 
\begin{itemize}
\item Average node degree scores of Protein\_181, Protein\_326, Protein\_923, Protein\_990 are 11.47, 9.22, 11.13, and 13.50 respectively; 

\item Average node betweenness scores of Protein\_181, Protein\_326, Protein\_923, Protein\_990 are 0.16, 0.13, 0.09, and 0.14 respectively. 
\end{itemize}
That is, these traditional graph characteristics are virtually indistinguishable. Furthemore, the state-of-the-art baselines (e.g., GraphCL, AD-GCL, and AutoGCL) {\bf do not} correctly classify these 4 proteins. For instance, the AutoGCL classifies Protein\_181, Protein\_326, and Protein\_990 in the same class, and the AD-GCL classifies Protein\_181, Protein\_326, and Protein\_923 in the same class. 

We now turn to the extended persistence, extract EPD of each network and compute pairwise Wasserstein distances between EPDs (shown as follows). 
\begin{itemize}
\item $W(\text{Protein}_{181}, \text{Protein}_{990}) = 2.768$

\item $W(\text{Protein}_{326}, \text{Protein}_{923}) = 2.578$

\item $W(\text{Protein}_{181}, \text{Protein}_{326}) = 5.184$

\item $W(\text{Protein}_{181}, \text{Protein}_{923}) = 4.549$

\item $W(\text{Protein}_{990}, \text{Protein}_{326}) = 5.713$

\item $W(\text{Protein}_{990}, \text{Protein}_{923}) = 4.887$

\end{itemize}
The results suggest that Protein\_181 and Protein\_990 are in the same class, and Protein\_326 and Protein\_923 also belong to the same class since they have smaller Wasserstein distances. In general, we find the Wasserstein distances between two EPDs of protein networks are always very high if those two protein networks do not belong to the same class and low, otherwise. These findings underline that persistence and topological invariance play essential roles in CL. Finally, in contrast to GraphCL, AD-GCL, and AutoGCL, TopoGCL correctly classifies all protein networks. 

\subsection{D.3. Algorithm Complexity}
The complexity to compute an 1-dimensional EPD is $\boldsymbol{O}(|\mathcal{V}||\mathcal{E}|)$. We also note that, while not officially published, the most recognized algorithm for EPD computation is quasilinear, i.e., $\boldsymbol{O}(\log(|\mathcal{V}|)|\mathcal{E}|)$ by using the data structure of mergeable trees~\cite{georgiadis2011data}.

\section{E. Additional Experiments}
\begin{table}[!htb]
\centering
%\caption{Ablation study.}
%\begin{minipage}{0.53\linewidth}
\setlength\tabcolsep{5pt}
\begin{tabular}{llccc}
\toprule
&\textbf{Architecture} & \textbf{Accuracy} (\%) \\
\midrule
\multirow{2}{*}{\textbf{MUTAG}}&{TopoGCL + EPI} &  {\bf 90.09$\pm$0.93}\\
&TopoGCL + PI & 89.78$\pm$1.33\\
\midrule
\multirow{2}{*}{\textbf{PTC\_MR}}&{TopoGCL + EPL} & {\bf 63.43$\pm$1.13} \\
&TopoGCL + PL & 62.78$\pm$0.87\\
\midrule
\multirow{2}{*}{\textbf{IMDB-B}}&{TopoGCL + EPI} & $^{***}${\bf 74.67$\pm$0.32} \\
&TopoGCL + PI & 70.55$\pm$0.19\\
\bottomrule
\end{tabular}
\caption{Ablation study of EPL and EPI.\label{ablation_study_1}}
\end{table}

\begin{table}[!htb]
\centering
\setlength\tabcolsep{3pt}
\begin{tabular}{llccc}
\toprule
&\textbf{Architecture} & \textbf{Accuracy} (\%) \\
\midrule
\multirow{2}{*}{\textbf{MUTAG}}&{TopoGCL} & $^{***}${\bf 90.09$\pm$0.93}\\
&TopoGCL W/o Topo. & 87.85$\pm$0.79\\
\midrule
\multirow{2}{*}{\textbf{PTC\_MR}}&{TopoGCL} & $^{***}${\bf 63.43$\pm$1.13} \\
&TopoGCL W/o Topo. & 61.62$\pm$1.10\\
\midrule
\multirow{2}{*}{\textbf{IMDB-B}}&{TopoGCL} & $^{***}${\bf 74.67$\pm$0.32} \\
&TopoGCL W/o Topo. & 71.31$\pm$0.36\\
\bottomrule
\end{tabular}
\caption{Ablation study of contrastive modes.\label{ablation_study_2}}
\end{table}

%%%

\end{document}